\documentclass[times,twocolumn,final]{elsarticle}

\usepackage{lineno,hyperref}

\usepackage{framed,multirow}

\usepackage{amssymb}
\usepackage{latexsym}

\usepackage{url}
\usepackage{xcolor}
\definecolor{newcolor}{rgb}{.8,.349,.1}

\usepackage{subfigure}

\usepackage{amsmath}
\usepackage{setspace}
\usepackage{multirow}
\usepackage{algorithm}
\usepackage{algorithmicx}
\usepackage{algpseudocode}
\modulolinenumbers[100]

\begin{document}

\begin{frontmatter}

\title{Outlier Detection Using a Novel method: Quantum Clustering}


\author[1]{Ding Liu\corref{mycorrespondingauthor}}
\cortext[mycorrespondingauthor]{Corresponding author}
\ead{liuding@tiangong.edu.cn}

\author[2]{Hui Li}

\address[1]{School of computer science and technology, Tiangong University, Tianjin, China}
\address[2]{Shanghai Library, Shanghai, China}

\begin{abstract}
We propose a new assumption in outlier detection: Normal data instances are commonly located in the area that there is hardly any fluctuation on data density, while outliers are often appeared in the area that there is violent fluctuation on data density. And based on this hypothesis, we apply a novel density-based approach to unsupervised outlier detection. This approach, called Quantum Clustering (QC), deals with unlabeled data processing and constructs a potential function to find the centroids of clusters and the outliers. The experiments show that the potential function could clearly find the hidden outliers in data points effectively. Besides, by using QC, we could find more subtle outliers by adjusting the parameter $\sigma$. Moreover, our approach is also evaluated on two datasets (Air Quality Detection and Darwin Correspondence Project) from two different research areas, and the results show the wide applicability of our method.
\end{abstract}

\begin{keyword}
%\texttt{Quantum Clustering}\sep Outlier detection
Quantum Clustering \sep Outlier detection
%\MSC[2010] 00-01\sep  99-00
\end{keyword}

\end{frontmatter}

\linenumbers

\section{Introduction}
\label{intro}
Outlier detection, also know as Anomaly Detection, is a process of finding non-conforming data or pattern in a dataset, and has been subjected to intensive examination \citep{chandola2009anomaly,hodge2004survey,chandola2012anomaly}. It is widely used in a variety of applications such as fraud detection for credit cards \citep{van2015apate}, insurance or health care \citep{aggarwal2005abnormality} and intrusion detection for cyber-security \citep{mahoney2002learning,bhuyan2014network}. There are several fundamental approaches to the problem of outlier detection such as classification, clustering, Nearest Neighbor based method, statistical method, spectral method \citep{chandola2009anomaly}, even deep learning based method \citep{chalapathy2019deep}. Different from these fundamental approaches, our approach in this paper, is a novel unsupervised Quantum Clustering technique \citep{horn2001algorithm}. It shows a great efficiency and fits the outlier detection very well.

As we mentioned above, many approaches have been developed to address the outlier detection problem and most of the existing approaches are proposed to solve the problem from specific aspects. For instance, the idea of classification-based approach is to build a predictive model to differentiate normal data and anomaly data. And it is normally considered as a supervised model. However, obtaining accurate and sufficient labels, especially for the outliers is usually a challenging task. While the clustering-based model does not require labeled data, and thus is applied to more places. It can be considered as a kind of typical unsupervised model. Inspired by this idea, we extend the application of QC to outlier detection, and we find this algorithm could reveal the subtle fluctuation on data density and help us to find the potential outliers. Different from some existing concepts, QC based outlier detection method is based on a new following assumption:

\vspace{0.08in}
\emph{Assumption 1: Normal data instances are commonly located in the area that there is hardly any fluctuation on data density, while outliers are often appeared in the area that there is violent fluctuation on data density.}
\vspace{0.08in}
\\Actually, this is exactly the key thinking of employing QC to outlier detection. Some other related existing assumptions will be discussed in section \ref{sec:Related works}.

Quantum Clustering emerged as a new approach of density-based clustering. It derives from the principles of quantum mechanics and finds clusters by building the so-called potential function. This is the essential part of QC and why it outperforms the conventional technique since the potential function is better at revealing the underlying structure of the data. In our previous work \citep{liu2016analyzing}, QC already has been applied to text analysis and achieved good performance. So based on these, in this paper, we start to employ QC to the outlier detection. The details regarding the advantages and the limitations of QC were also provided. Moreover, we employed it to two experimental application i.e. Darwin correspondence project and air quality detection. The experimental results show us the QC could be a well choice for the outlier detection.

\section{Related work}
\label{sec:Related works}
Clustering-based method and the density-based method are two primary techniques commonly used in static outlier detection. According to previous studies, clustering-based approach can be grouped into three categories based on three different assumptions \citep{chandola2009anomaly}:

\vspace{0.08in}
\emph{Assumption 2: Normal data instances belong to a cluster in the data, while outliers either do not belong to any cluster.}

\vspace{0.08in}
\emph{Assumption 3: Normal data instances lie close to their closest cluster centroid, while outliers are far away from their closest cluster centroid.}

\vspace{0.08in}
\emph{Assumption 4: Normal data instances belong to large and dense clusters, while outliers either belong to small or sparse clusters.}

Many famous algorithms used in outlier detection are based on at least one of these assumptions, such as DBSCAN \citep{ester1996density} (assumption 2), FindOut \citep{yu2002findout} (assumption 2), K-means \citep{hartigan1979algorithm} (assumption 3), SOM \citep{kohonen1998self} (assumption 3), and FindCBLOF \citep{he2003discovering} (assumption 4).

Besides, there is also an assumption as the basis for density-based approaches.\citep{chandola2009anomaly}:

\vspace{0.08in}
\emph{Assumption 5: Normal data instances occur in dense neighborhoods, while anomalies occur far from their closest neighbors.}

Among all density-based approaches, Nearest Neighbor (NN) labels the outlier by estimating data density and is widely used in clustering. Moreover, researchers have proposed LOF and COF algorithms to address the error in NN, which is caused by a specific data distribution \citep{breunig2000lof}. We conclude all these algorithms and their corresponding assumptions in Table.\ref{tab:1}

\begin{table}
\centering
\begin{spacing}{1.2}
\caption{Existing algorithms and their corresponding assumptions }
\label{tab:1}       % Give a unique label
\begin{tabular}{ll}
\hline\noalign{\smallskip}
Assumptions & Algorithms\\
\noalign{\smallskip}\hline\noalign{\smallskip}
2  & DBSCAN, FindOut              \\
3  & Kmeans, SOM                  \\
4  & FindCBLOF                    \\
5  & NN, LOF, COF                 \\
\noalign{\smallskip}\hline
\end{tabular}
\end{spacing}
\end{table}

Our algorithm QC, is a hybrid method combined with clustering and density-based approach. It is based on assumption 1, 4 and 5. As it has advantages of both clustering and density-based approach, it could also resolve the specific problem such as we present in Fig.\ref{fig:1:d}, which KNN could not handle.

%Besides the assumption 1 we have mentioned, QC could also handle the common case relies on the assumption 4 and the assumption 5. Therefore, it is a hybrid technique composed by the thinking of clustering and the density-based method. It combines the metric of both of them, and could also resolve the specific problem which KNN could not handle. We presented it in Fig.\ref{fig:1:d}. And the details of QC will be discussed in Sec.\ref{sec:method}.

\section{Method}
\label{sec:method}
The detailed explanation of the Quantum Clustering could be found at the previous works \citep{horn2001method,horn2001algorithm,nasios2007kernel,liu2016analyzing}. But for self-containing, we need to contain some significant parts including principle, method and algorithm in this section.

\subsection{Algorithm}
\label{sec:Algorithm}
The algorithm of Quantum Clustering is proposed based on the Schr\"{o}dinger equation, which is considered as the foundation in quantum mechanics. In the principle of quantum theory, this equation describes the dynamic behavior of a microscopic physical system. Our algorithm QC can also describe data clustering process by this equation and in this situation, the algorithm could be deduced by the following steps.
The time-independent Schr\"{o}dinger equation is given as Eq. (\ref{equ:1}) \citep{feynman1965lectures}

\begin{equation}
\label{equ:1}
H\psi\left(x \right)=\left(-\frac{\hbar^{2}}{2m}\nabla^{2} +v(x)\right)\psi \left(x \right)
                    =E\psi \left(x \right)
\end{equation}

\noindent
where H is the Hamiltonian operator. The \emph{$\psi(x)$} refers to the so-called wave function and \emph{v(x)} denotes the potential function. By considering the wave function \emph{$\psi(x)$} as the known condition in the Schr\"{o}dinger equation, we aim to determine the potential \emph{v(x)} in Quantum Clustering, which characterizes the probability density function of the input data\citep{nasios2007kernel}. Therefore, the essential part of the algorithm is to calculate \emph{v(x)} by using Schr\"{o}dinger equation. Given Gaussian kernel as Eq. (\ref{equ:2}) for the wave function, and $m =\hbar^2/\sigma^2 $, where $\sigma$ denotes the width parameter.

\begin{equation}
\label{equ:2}
\psi \left(x\right)=\sum_{i}e^{-\left(x-{x}_{i} \right)^{2}/2\sigma ^{2}}
\end{equation}

Thus, the potential function \emph{v(x)} could be solved as:

\begin{equation}
\label{equ:7}
\begin{aligned}
v(x)&=E+\frac{\sum_{i}(e^\frac{-(x-x_{i})^2}{2\sigma^2} \cdot\frac{(x-x_{i})^2}{2\sigma^2}-e^\frac{-(x-x_{i})^2}{2\sigma^2}\cdot\frac{1}{2})}{\sum_{i}e^\frac{-(x-x_{i})^2}{2\sigma^2}}\\
&=E-\frac{1}{2}+\frac{1}{2\sigma^2\psi(x)}\sum_{i}(x-{x}_{i})^2e^\frac{-(x-x_{i})^2}{2\sigma^2}\\
&\approx \frac{1}{2\sigma^2\psi(x)}\sum_{i}(x-{x}_{i})^2e^\frac{-(x-x_{i})^2}{2\sigma^2}
\end{aligned}
\end{equation}

\begin{algorithm}
\label{Alg:1}
\caption{Calculating the Potential function for each data point}
    \begin{algorithmic}[1]
        \Require $data:$ data points,
        \par $\sigma:$ width parameter,
        \par $n:$ the number of data points
             \Function {POTENTIAL}{$data(i)$}
             \State $sum1 \gets 0$
             \State $sum2 \gets 0$
               \For{$j = 1 \to n$}
                    \State $dist(j)\gets \sqrt{(data(i)-data(j))^2}$;
                    \State $sum1 \gets sum1+dist(j)^2\cdot e^{-dist(j)^2/2\sigma^2}$
                    \State $sum2 \gets sum2+e^{-dist(j)^2/2\sigma^2}$
               \EndFor
               \State $v(data(i)) \gets \frac{1}{2\sigma^2}\cdot \frac{sum1}{sum2}$
               \State \Return {$v(data(i))$}
            \EndFunction

\end{algorithmic}
\end{algorithm}

Then, some classic optimization approaches can be utilized to deduce the clustering allocation. In our study, we employed the classic BFGS( Broyden-Fletcher-Goldfarb-Shanno) algorithm, which is a kind of Quasi-Newton Methods (Broyden, 1970) to address the problem. The details of BFGS could be found in many previous publications(e.g., \citep{lewis2009nonsmooth,byrd1995limited}). At last, according to the assumption 1 we mentioned before, we address the issue by counting the amount of data each cluster contains. Algorithm 1 and 2 list the pseudocode of the calculation of potential function for each data point and quantum clustering separately. Hence, the time complexity of QC derived from two parts, i.e. computing potential function and the optimization algorithm. The time complexity of computing potential function is $O(n^2)$, which $n$ denotes the number of data points. And the counterpart of optimization depends on the algorithm we adopted. In our study, as we mentioned, we used the classical BFGS algorithm.

\begin{algorithm}
\label{Alg:2}
\caption{Quantum clustering for outlier detection}
    \begin{algorithmic}[1]

        \Require $data:$ data points,
        \par $n:$ the number of data points
            \State $\sigma \gets$ Initial value;
            \State $k \gets$ Initialise the threshold;
            \For{$i = 1 \to n$}
                    \State Input each data(i);
                    \State Optimize the  POTENTIAL$(data(i))$ by BFGS;
                    \State Assign the clustering labels to each data;
            \EndFor
            \State $m \gets$ number of clusters;
            \For{$j = 1 \to m$}
                    \If {size of cluster(j)$<k$}
                    \State Label all data in cluster(j) as outliers;
                    \EndIf
            \EndFor

\end{algorithmic}
\end{algorithm}

\subsection{Performance}
\label{sec:Performance}
We tested the performance of QC in outlier detection by using several typical artificial datasets. These tests reveals the advantages of QC, and based on these, we understand which types of dataset is suitable for applying QC.

\begin{figure*}[htbp]
  \centering
\subfigure[]{

    \label{fig:1:a} %% label for first subfigure
    \includegraphics[width=2in,height=1.6in]{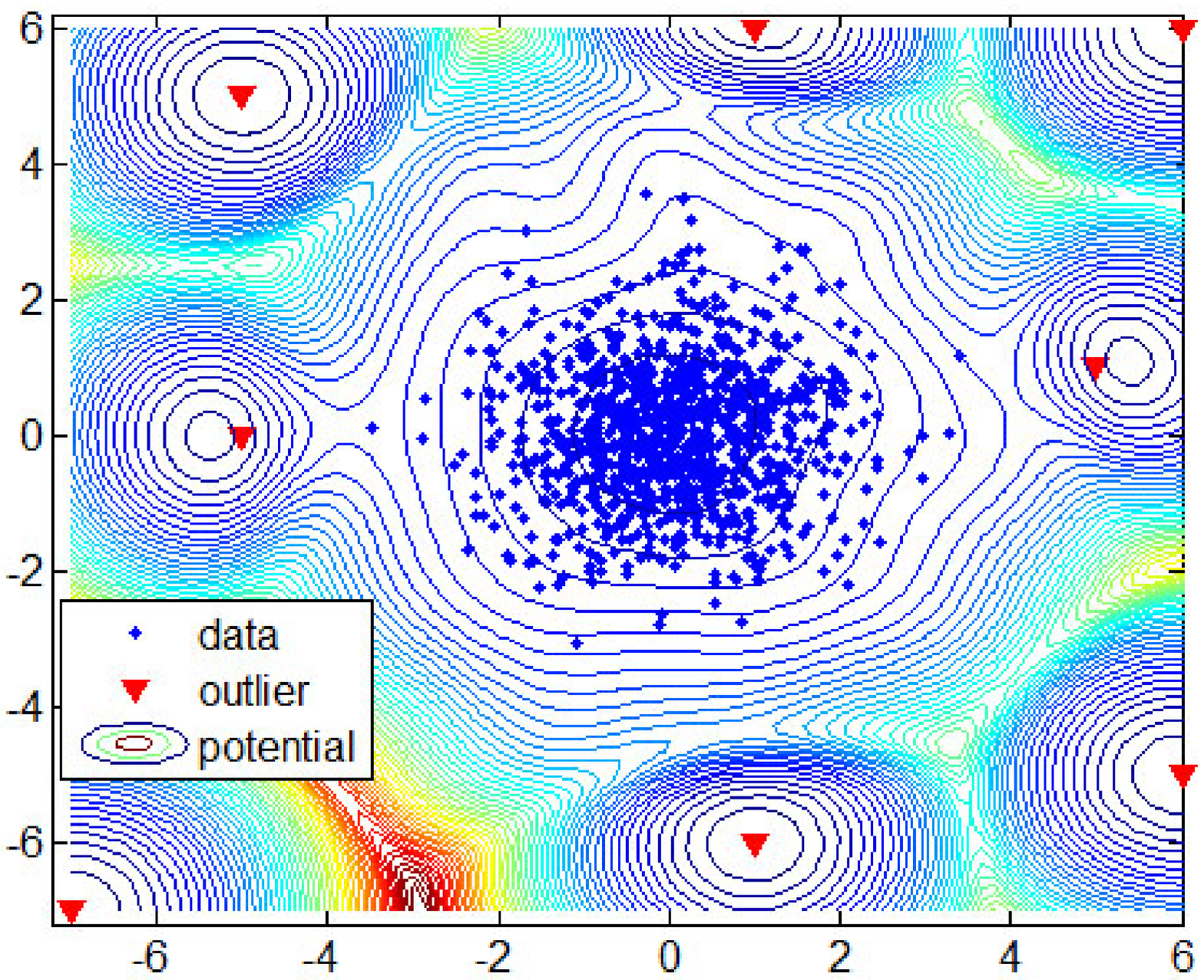}}
    %\hspace{1in}
  \subfigure[]{
    \label{fig:1:b} %% label for second subfigure
    \includegraphics[width=2in,height=1.6in]{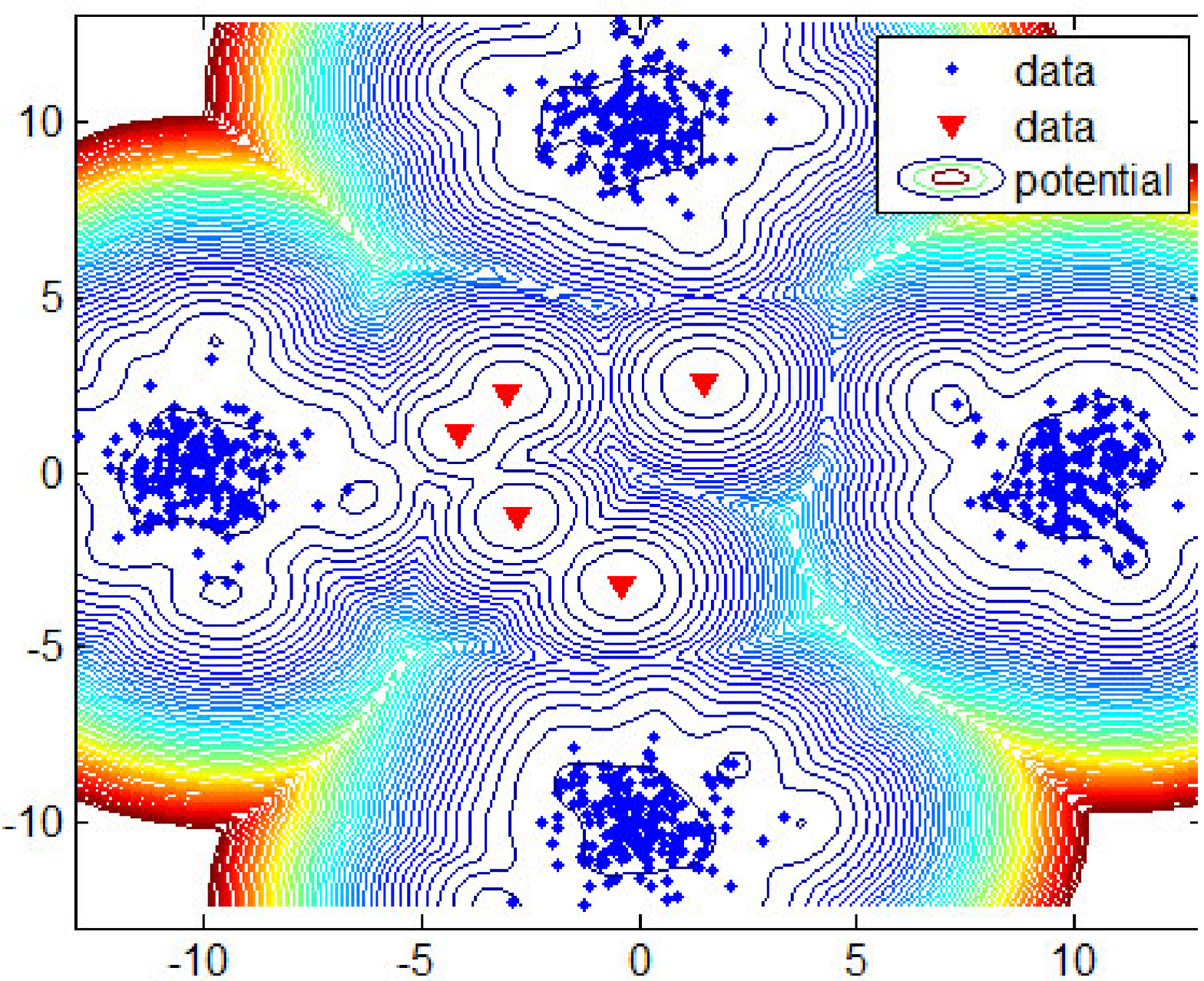}}
    %\hspace{1in}
  \centering
  \subfigure[]{
    \label{fig:1:c} %% label for first subfigure
    \includegraphics[width=2in,height=1.6in]{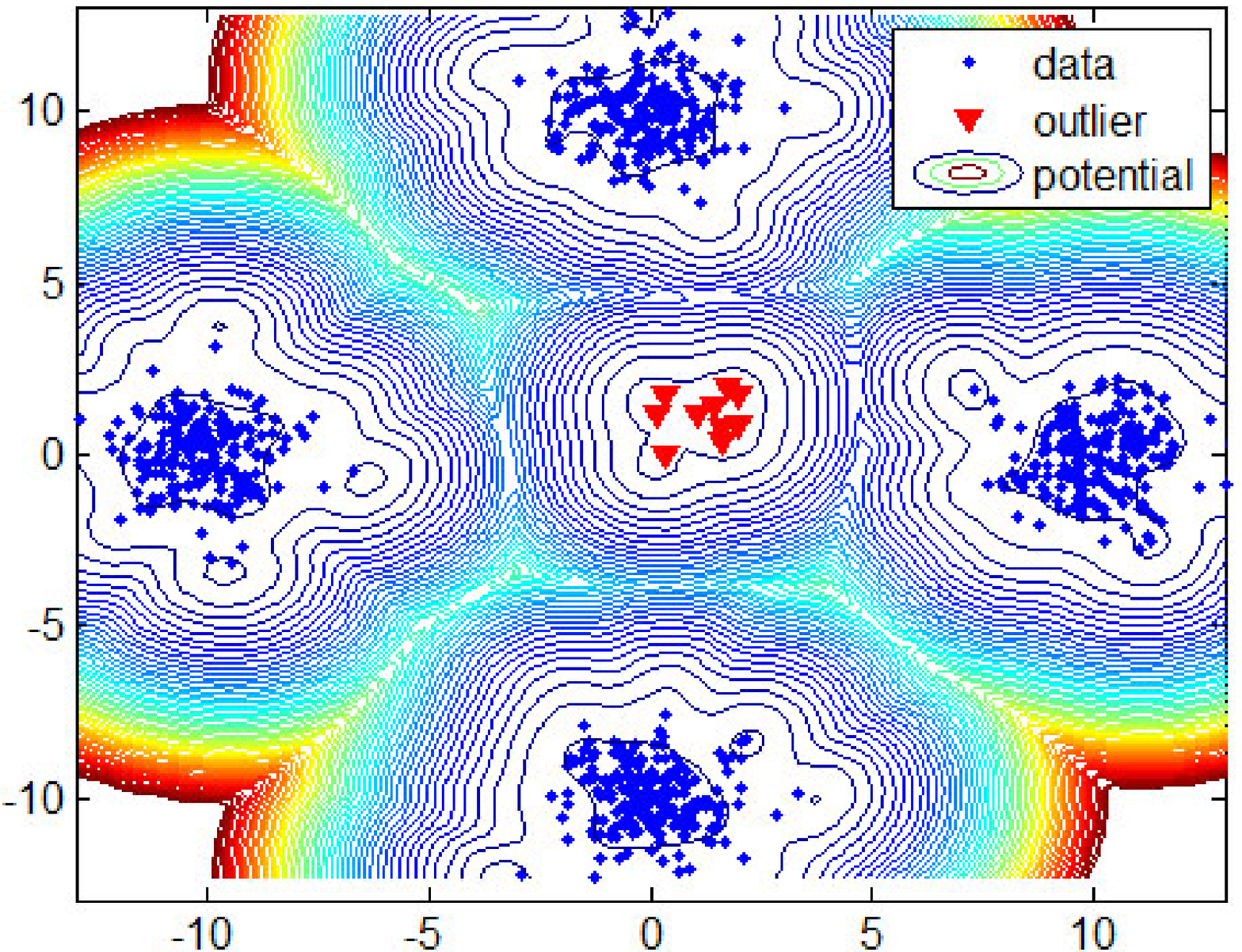}}

    \subfigure[]{
    \label{fig:1:d} %% label for first subfigure
    \includegraphics[width=2in,height=1.6in]{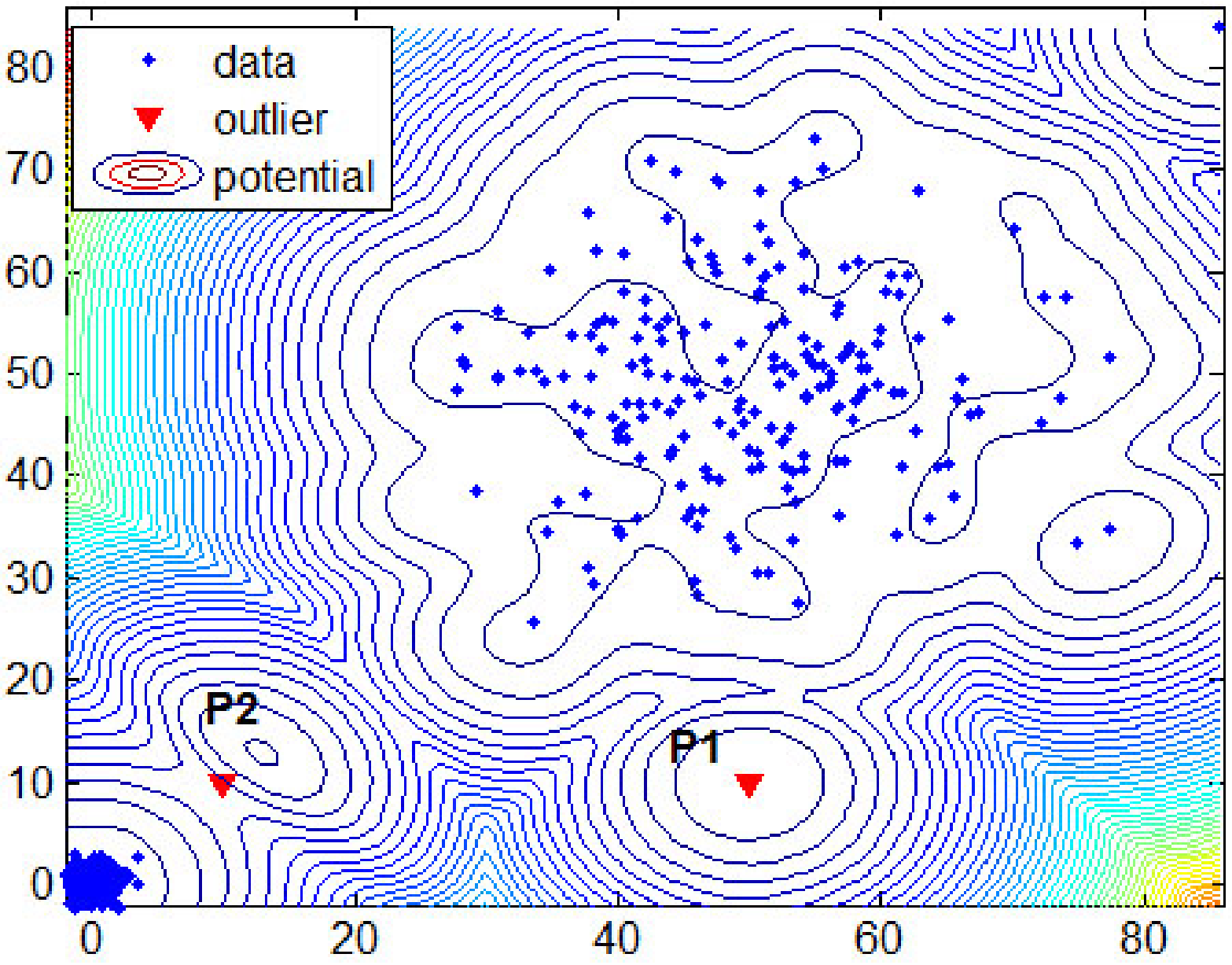}}
    %\hspace{1in}
  \subfigure[]{
    \label{fig:1:e} %% label for second subfigure
    \includegraphics[width=2in,height=1.6in]{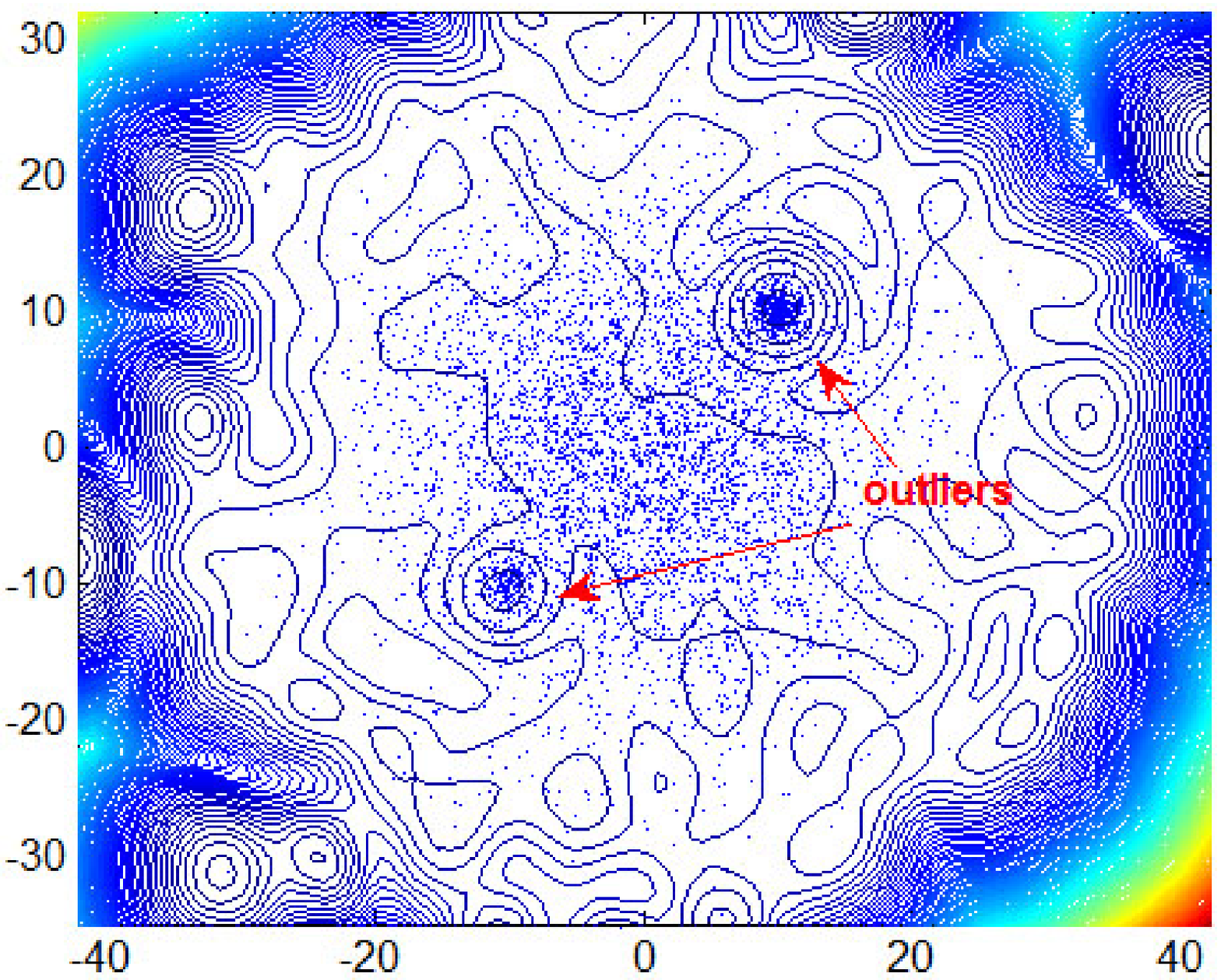}}
    %\hspace{1in}
  \centering
  \subfigure[]{
    \label{fig:1:f} %% label for first subfigure
    \includegraphics[width=2in,height=1.6in]{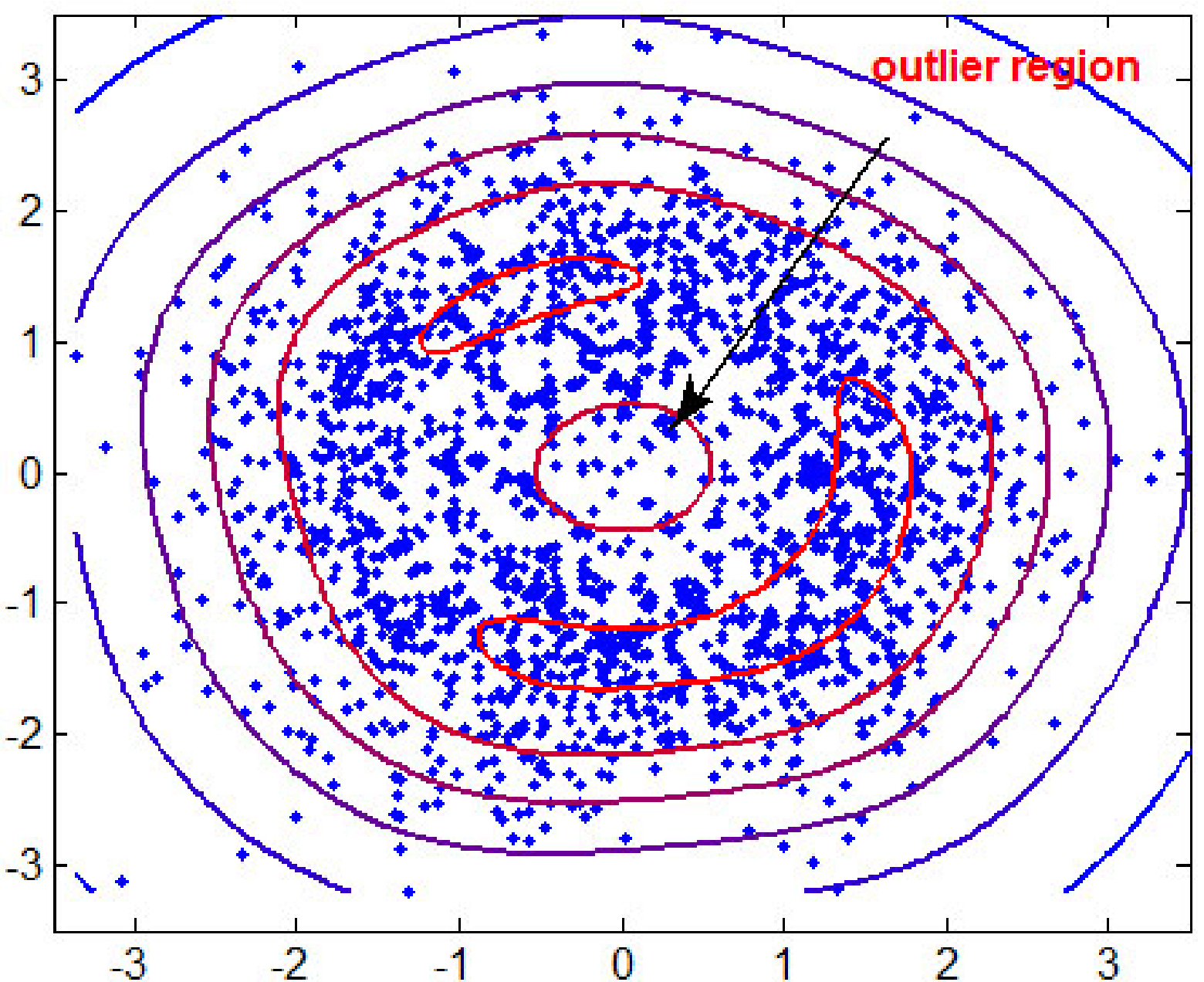}}
  %\hspace{1in}
  \caption{Experiments on artificial datasets, and the contour line of potentials are also displayed; (a) Normal data surrounded by separated outliers; (b) Separated outliers surrounded by normal data; (c) Clustered outliers surrounded by normal data; (d) A specific data distribution in which the outlier P1 is difficult to be detected by the KNN; (e) A specific data distribution in which a main data cluster is created following the Gaussian distribution. And on the edage of this area, two small high-density clusters is produced, thus these two small ones formed violent density fluctuation in the main cluster; (f) A specific data distribution in which a small sparse data region is surrounded by high-density data. Thus, the data in the sparse region are the outliers. We should emphasize that the potential function is replaced by the inverse of it;}
  \label{fig:1} %% label for entire figure
\end{figure*}

Fig.\ref{fig:1} displays artificial 2-D data distributions for outlier detection, which has been already discussed in Ref.\citep{chandola2009anomaly}. These are all typical data distribution for outlier detection, and show us some specific cases. We find QC detect outliers very easily in all these cases. Moreover, if we apply KNN (kth Nearest neighbor) algorithm in the data distribution displayed by panel \ref{fig:1:d}, the data point ``P1'' will not be recognized correctly as outlier. But if we use QC algorithm, the outliers could be detected easily and completely.

In the Fig. \ref{fig:1:e} and Fig. \ref{fig:1:f}, we presented two specific cases that show violent density fluctuation clearly. In Fig. \ref{fig:1:e}, we generated a main data cluster following the Gaussian distribution. And on the edges of this area, we also generated two small high-density clusters, and these two small ones formed violent density fluctuations in the main cluster. As expected, QC also succeed in detecting these outliers, which is really hard to be solved by other clustering methods such as K-means or DBSCAN. In contrast, in Fig. \ref{fig:1:f}, we generated a small sparse data region surrounded by high-density data points. To our knowledge, this unfrequent case was hardly investigated. But we think the data in the sparse region should also be considered as outliers, and it could be addressed by QC. We should mention that we use the inverse of the potential function instead of the original one.

\subsection{Parameter estimation}
\begin{figure*}[htbp]
  \centering
\subfigure[]{

    \label{fig:2:a} %% label for first subfigure
    \includegraphics[width=2in,height=1.6in]{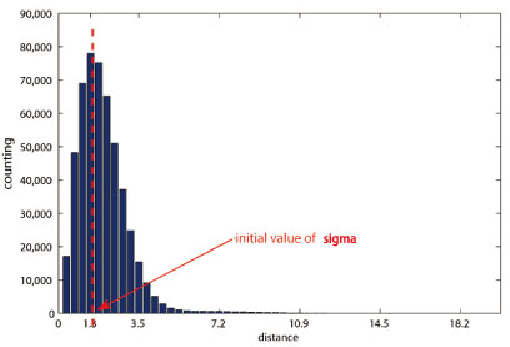}}
    %\hspace{1in}
  \subfigure[]{
    \label{fig:2:b} %% label for second subfigure
    \includegraphics[width=2in,height=1.6in]{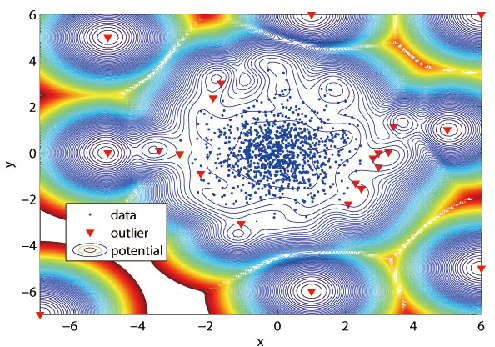}}
    %\hspace{1in}
  \centering
  \subfigure[]{
    \label{fig:2:c} %% label for first subfigure
    \includegraphics[width=2in,height=1.6in]{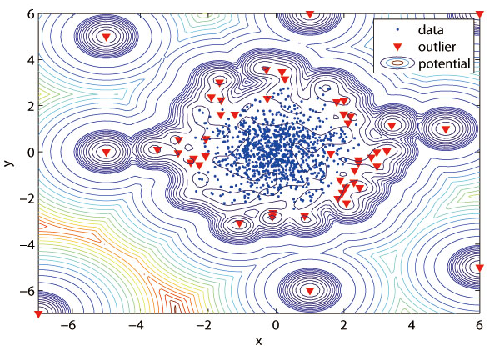}}

  %\hspace{1in}
  \caption{Parameter estimation; (a) Histogram of the distance; (b) Experimental result with $\sigma=0.5$; (c) Experimental result with $\sigma=0.3$ ; }
  \label{fig:2} %% label for entire figure
\end{figure*}

The algorithm is determined by the only width parameter $\sigma$. In our previous work \citep{liu2016analyzing}, we utilized the Pattern Search method to find an optimal clustering result. However, for outlier detection task, there is no need to find an optimal $\sigma$ because there is no optimal clustering result. Instead, we need to find an appropriate value of $\sigma$ that could distinguish the normal data and the outliers clearly.
As we showed in Fig.\ref{fig:2:a}, we first calculate the distance between each data point. And then, we select the value of distance which account for the largest portion as the initial value of $\sigma$. According to our experience, this value could distinguish between the normal data and the outliers clearly.

Furthermore, we could find more subtle outliers by decreasing the value of $\sigma$. Such as the case in Fig.\ref{fig:1:a}, if we decrease the value of $\sigma$ to 0.5, data points on the margin of the main cluster are recognised as outliers (Fig.\ref{fig:2:b}). Furthermore, more outliers emerge if we continue to decrease the value of $\sigma$ to 0.3 (Fig.\ref{fig:2:c}).

\section{Experiment \& Application}
\label{sec:Application}
We selected two applicable research area, i.e., history study and air quality detection, to demonstrate how the Quantum Clustering works in realistic task. At the first, it is employed to the Darwin Correspondence Dataset. And the second, we used it to analyze the air quality detection.

\subsection{Darwin Correspondence Project}
\label{sec:Darwin}

Charles Robert Darwin (1809-1882) is famous for his contributions to evolutionary theory. During his lifetime, Darwin used letters to exchange information and academic ideas with his friends, family and individuals who are helpful for his research.  These letters provide an access to the 19th century academic circles in science, culture and religion.
The researchers of the Darwin Correspondence Project based in Cambridge University Library collect, transcribe and publish most letters written by or to Darwin.  The dataset used in this paper is a subset of Darwin¡¯s letters spanning from 1821 to 1882. It consists of approximately 14,972 letters and 2,097 correspondents (i.e. sender or receiver).  This dataset is used to evaluate the proposed algorithms and validate the effectiveness and efficiency of proposed models.Such data is permitted by Darwin Correspondence Project to be used in this paper. And it is available at: https://www.darwinproject.ac.uk/

\begin{figure}
  \centering
    \includegraphics[width=3in,height=2in]{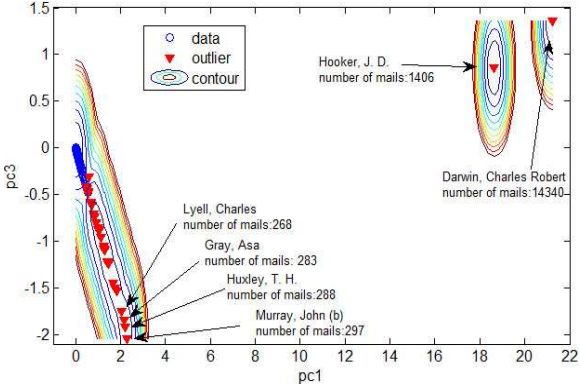}
    %\hspace{1in}

  \caption{The experimental results on Darwin Correspondence Project. Several correspondents who communicated with C.R.Darwin most frequently are marked on the figure with $\sigma=0.5$.}.
  \label{fig:3} %% label for entire figure
\end{figure}

And we use this dataset to demonstrated the performance of QC. First, we built a vector space model to describe the relationship between each correspondent by counting the number of letters of each person. And then, the original dataset was mapped to a 2-D feature space by two principle components which cover 87\% variance of the original data. The experimental results are presented in Fig.\ref{fig:3}. We find that all data are separated very clearly. The two ones in top right corner correspond to Charles Robert Darwin himself, and Joseph Dalton Hooker. As Mr. Hooker had the largest number of letters exchanged with Darwin, he seems to be in close relation with Darwin during that time. Besides, other persons who had frequent letter exchanges are also separated respectively. QC recognized all of them as the outliers and picked out them successfully. In comparison, most of the correspondents who have just several letters with Darwin are put together in one group.

\subsection{Air quality detection}
\label{sec:Air}
This dataset we used is a benchmark dataset which from the UCI Machine Learning Repository \citep{Lichman:2013,de2008field}. It contains 9358 instances from an Air Quality Chemical Multisensor Device. The device is located on the field in a significantly polluted area within an Italian city. Data were recorded from March 2004 to February 2005 representing the longest freely available recordings of on field deployed air quality chemical sensor devices responses \citep{de2008field}. Therefore, we investigated the non-conforming air quality periods by using QC to detect the outliers. This dataset consist of 15 attributes, which displayed in Table.\ref{tab:2}

\begin{table*}
\centering
\begin{spacing}{0.8}
\caption{Attributes of the Air quality detection dataset}
\label{tab:2}       % Give a unique label
\begin{tabular}{ll}
\hline\noalign{\smallskip}
NO. & Attribute description\\
\noalign{\smallskip}\hline\noalign{\smallskip}
0  & Date              \\
1  & Time	(HH.MM.SS) \\
2  & True hourly averaged concentration CO in $mg/m^3$ (reference analyzer)       \\
3  & PT08.S1 (tin oxide) hourly averaged sensor response (nominally CO targeted)	                   \\
4  & True hourly averaged overall Non Metanic HydroCarbons concentration in $microg/m^3$ (reference analyzer)              \\
5  & True hourly averaged Benzene concentration in $microg/m^3$ (reference analyzer) \\
6  & PT08.S2 (titania) hourly averaged sensor response (nominally NMHC targeted)	      \\
7  & True hourly averaged NOx concentration in ppb (reference analyzer)                   \\
8  & PT08.S3 (tungsten oxide) hourly averaged sensor response (nominally NOx targeted)               \\
9  & True hourly averaged NO2 concentration in $microg/m^3$ (reference analyzer)	 \\
10 & PT08.S4 (tungsten oxide) hourly averaged sensor response (nominally NO2 targeted)	      \\
11 & PT08.S5 (indium oxide) hourly averaged sensor response (nominally O3 targeted)                     \\
12 & Temperature \\
13 & Relative Humidity (\%)       \\
14 & AH Absolute Humidity                    \\
\noalign{\smallskip}\hline
\end{tabular}
\end{spacing}
\end{table*}

\begin{figure}
  \centering
\subfigure[]{

    \label{fig:4:a} %% label for first subfigure
    \includegraphics[width=1.5in,height=1.3in]{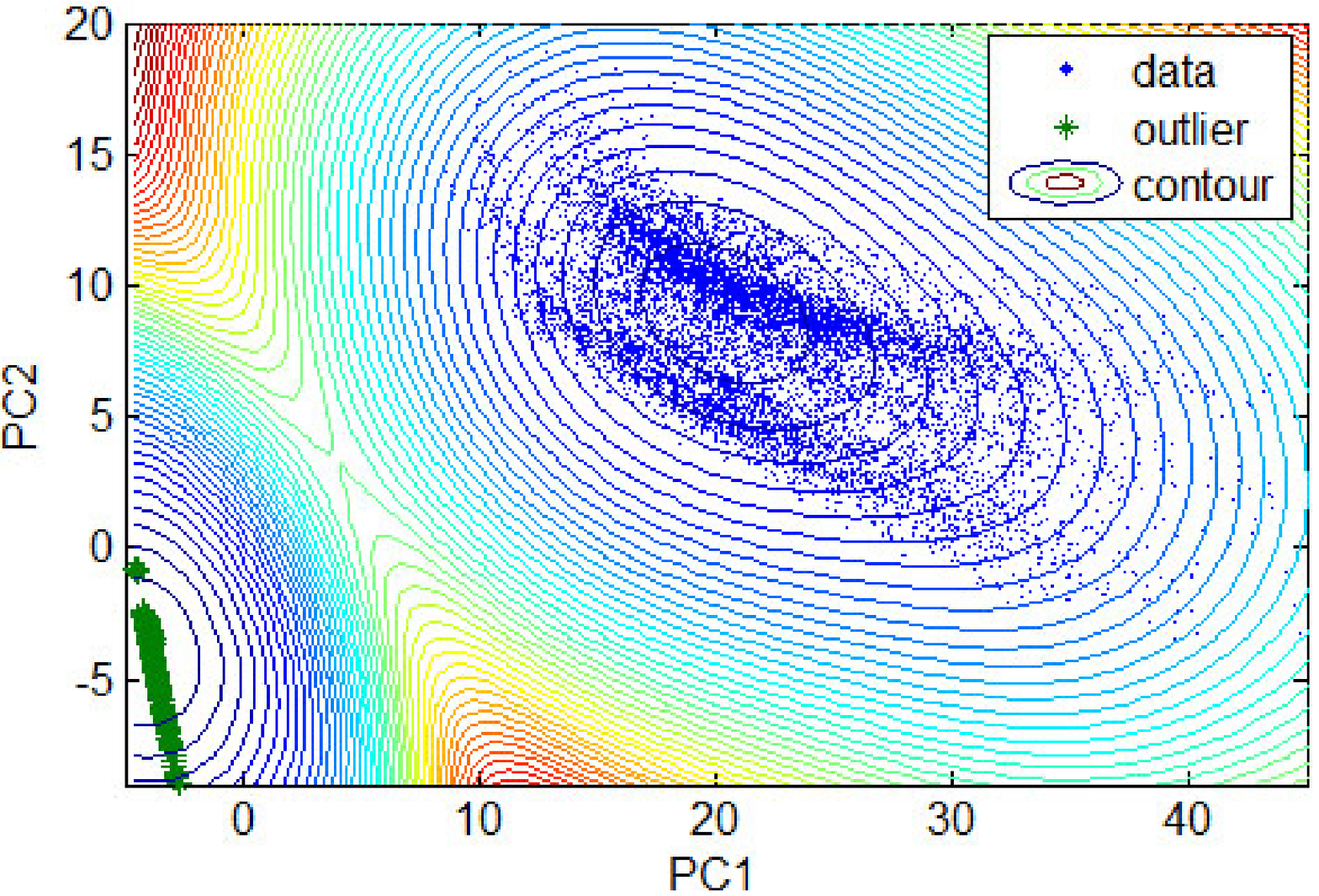}}
    %\hspace{1in}
  \subfigure[]{
    \label{fig:4:b} %% label for second subfigure
    \includegraphics[width=1.5in,height=1.28in]{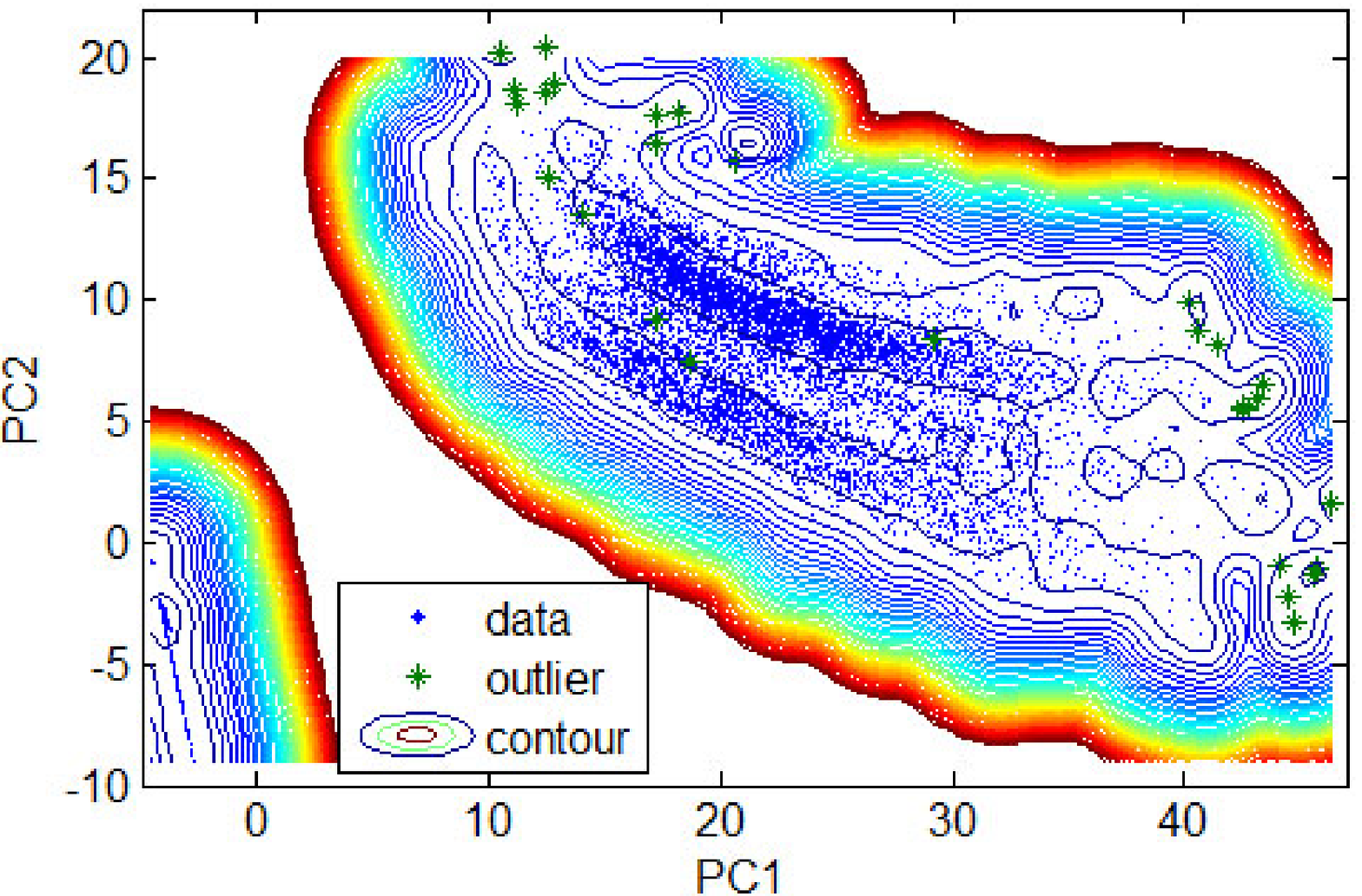}}
    %\hspace{1in}
  \caption{Data distribution of the air quality dataset; (a) $\sigma=6$; (b) $\sigma=1$}.
  \label{fig:4} %% label for entire figure
\end{figure}

\begin{figure*}
  \centering
\subfigure[]{

    \label{fig:5:a} %% label for first subfigure
    \includegraphics[width=1.8in,height=1.1in]{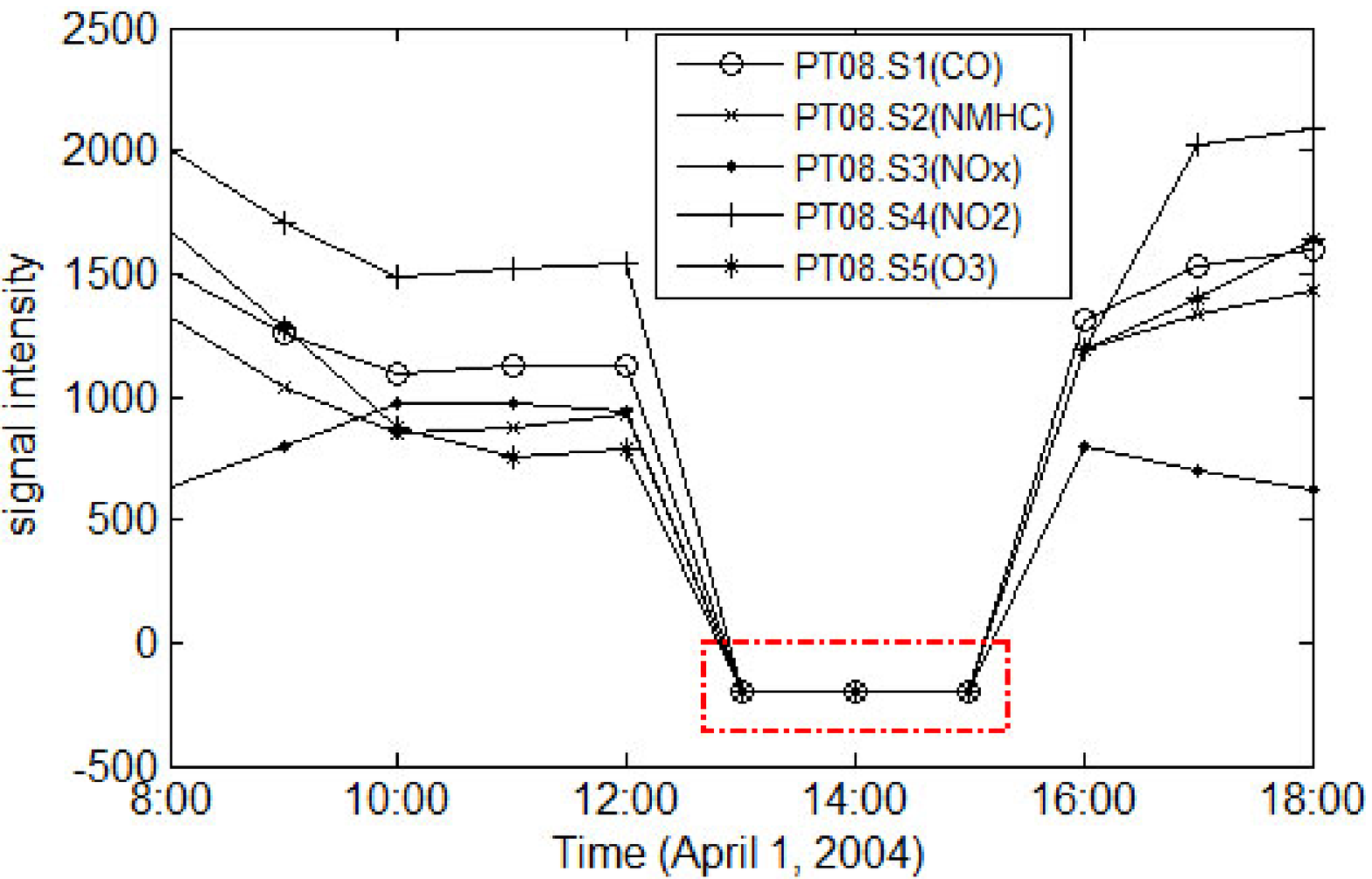}}
    %\hspace{1in}
  \subfigure[]{
    \label{fig:5:b} %% label for second subfigure
    \includegraphics[width=1.8in,height=1.1in]{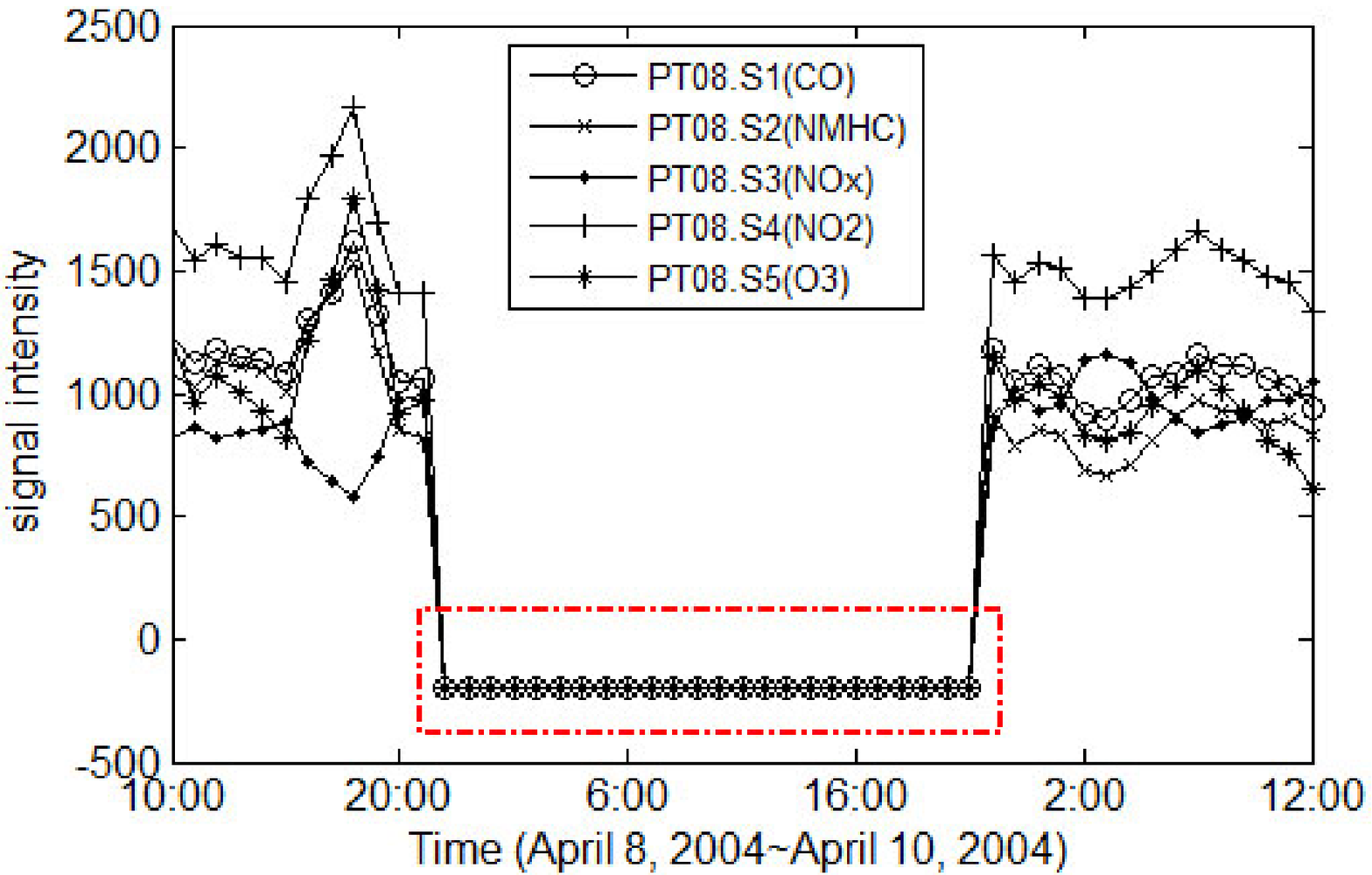}}
    %\hspace{1in}
  \subfigure[]{
    \label{fig:5:c} %% label for second subfigure
    \includegraphics[width=1.8in,height=1.1in]{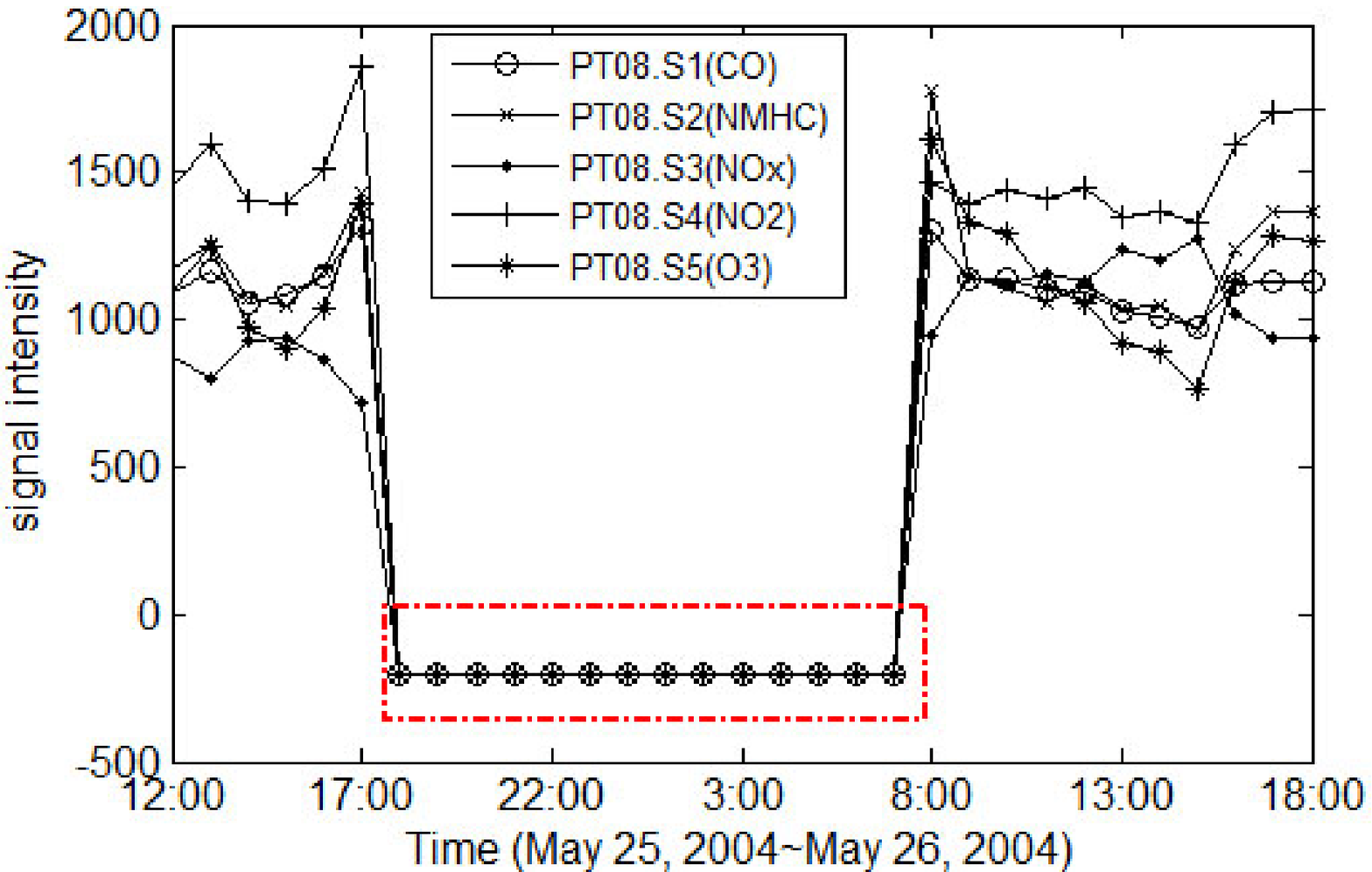}}
    %\hspace{1in}

  \caption{Original curve corresponding to part of outliers in Fig.\ref{fig:4:a}; (a) April 1, 2004; (b) April 8, 2004 $\sim$ April 10, 2004; (c) May 25, 2004 $\sim$ May 26, 2004; }.
  \label{fig:5} %% label for entire figure
\end{figure*}

\begin{figure*}
  \centering
\subfigure[]{

    \label{fig:6:a} %% label for first subfigure
    \includegraphics[width=1.8in,height=1.1in]{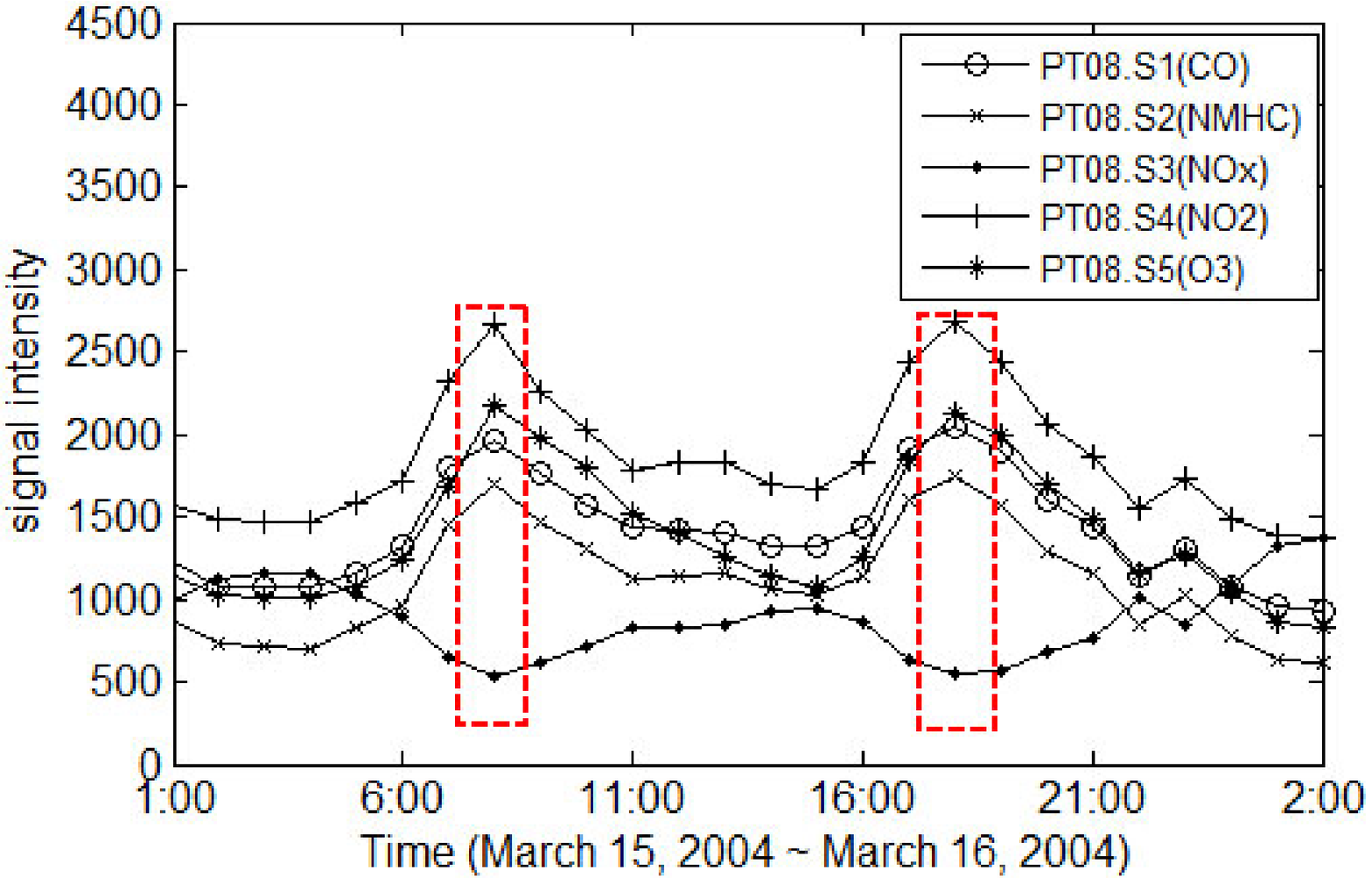}}
    %\hspace{1in}
  \subfigure[]{
    \label{fig:6:b} %% label for second subfigure
    \includegraphics[width=1.8in,height=1.1in]{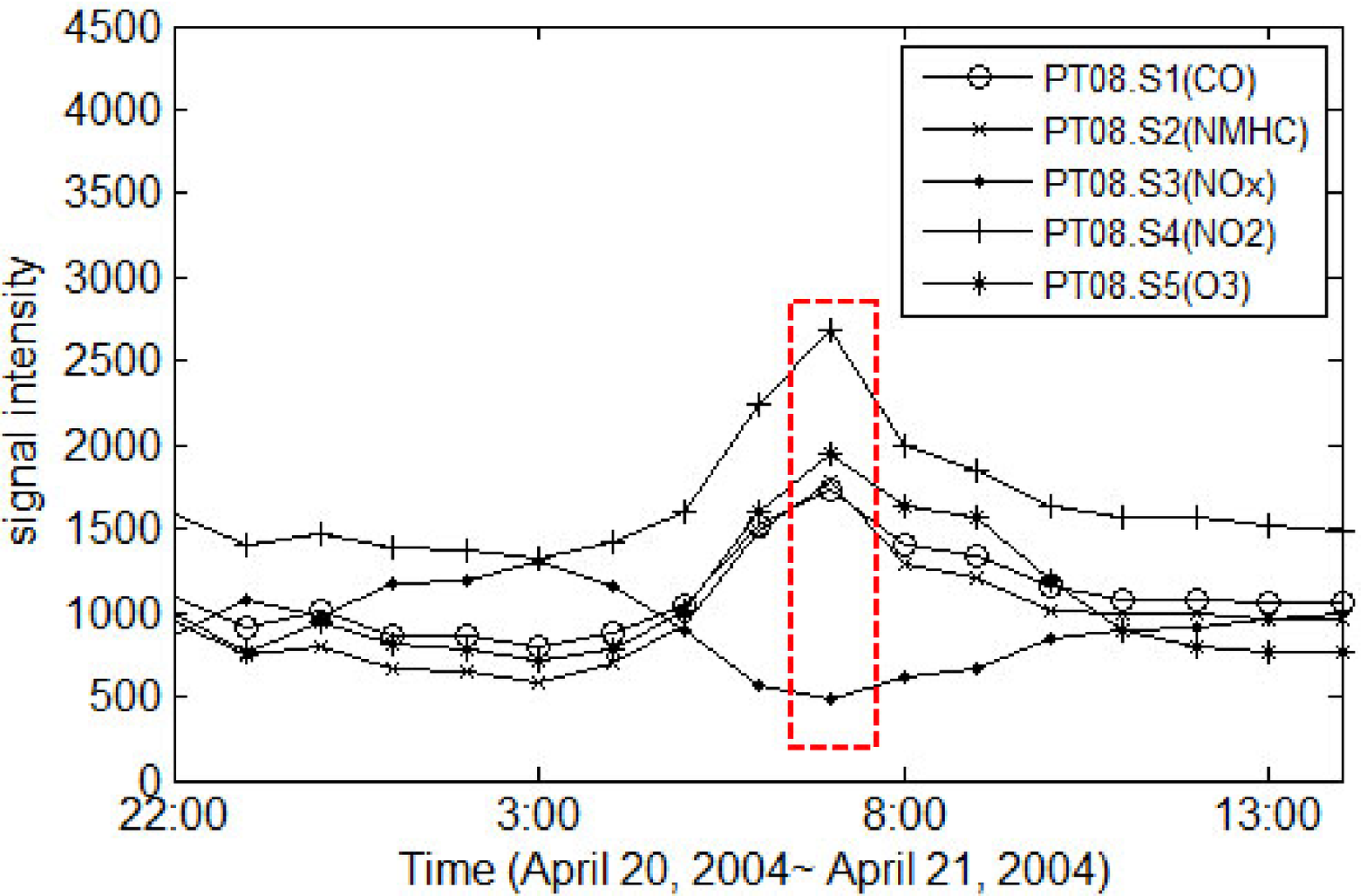}}
    %\hspace{1in}
  \subfigure[]{
    \label{fig:6:d} %% label for second subfigure
    \includegraphics[width=1.8in,height=1.1in]{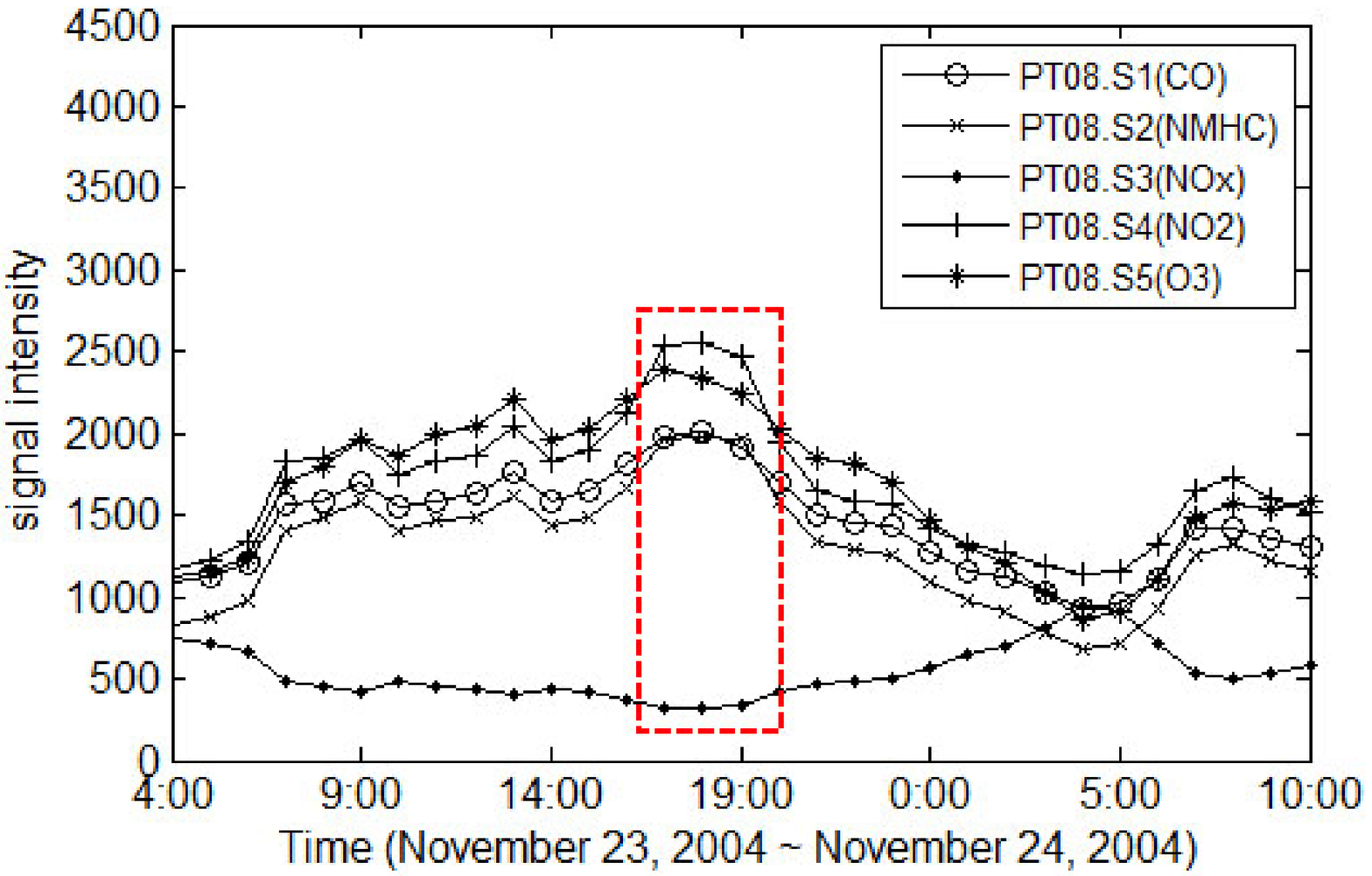}}
    %\hspace{1in}

  \caption{Original curve corresponding to part of outliers in Fig.\ref{fig:4:b}; (a) March 15, 2004 $\sim$ March 16, 2004; (b) April 20, 2004 $\sim$ April 21, 2004; (c) November 23, 2004 $\sim$ November 24, 2004;}.
  \label{fig:6} %% label for entire figure
\end{figure*}

First, PCA was also used to reduce the high dimension. And we selected top 2 principle components that contain 85\% variance to represent the original data. Afterwards, we perform QC to find the outliers. The 2-D data points, contour of the potential function and the experimental results were plotted in Fig.\ref{fig:4:a}. Obviously, QC easily detected the outliers that correspond to the small cluster when $\sigma$ equal to 6. Furthermore, we supposed that the outliers should indicate the abnormal air quality data points in the dataset. Therefore, we picked up some outliers and ploted the original data curve corresponding to these outliers in Fig.\ref{fig:5}. to verify this assumption. The outliers we detected correspond to the abnormal data points that have absolute low values in nearly every air quality indicator. Based on this, we could know when this polluted area has relative good air quality.

Moreover, we explored more patterns by decreasing $\sigma$. As presented in Fig.\ref{fig:4:b}, it is obvious that QC discover some other outliers around the main cluster. Besides, the corresponding curve is shown in Fig.\ref{fig:6} and the peak of the curve corresponds to the time points when air pollution is at its worst.

\section{Discussion}
\label{sec:Discussion}
Based upon the experiments discussed above, we make some conclusions that:

1.	We propose a new key assumption that:

%\vspace{0.08in}
\emph{Normal data instances are commonly located in the area that there is hardly any fluctuation on data density, while outliers are often appeared in the area that there is violent fluctuation on data density.}
%\vspace{0.08in}

This is the foundation of applying QC to outlier detection, and to our knowledge, it is probably that QC is the most powerful algorithm to find subtle variation of density of data. As we discussed in Ref.\citep{liu2016analyzing}, QC outperform conventional Parzen-window and DBSCAN.

2.	QC is considered as a flexible hybrid technique consisting of potential function (objective function) and optimization algorithm. Hence, QC could fit different cases by using different objective functions. Such as the case we presented in Fig.(\ref{fig:1:f}).

3.	QC is a kind of unsupervised method, and do not make assumptions regarding the data distribution. Thus, it is a type of data driven approach.

However, in addition to the significant advantages of QC, we also would like to discuss some limitations of it as follows:

1.	The time complexity of distance computing is $O(n^2)$, which $n$ denotes the number of input data points. It will be a big challenge when we try to apply the QC to some big data tasks since it need to traverse all input data. In our further study, we will try to narrow the circle of data points that involved in distance computing. That is, we will just involve the data points very close to the target point to compute the distance of target point. Definitely, it will reduce the time complexity to $O(m^2)$, which $m$ denotes the number of data points very close to the target point, and it is far less than $n$. Of course, it should be noted that this strategy will lead to a slight decrease of potential function accuracy.

2.	As an unsupervised-based approach, QC fits the static outlier detection very well. But it could not address the dynamic case. In other words, it can't be used as a method which could predict whether a new instance is an outlier or a normal data point, such as what some supervised-based methods can do.

\section*{Acknowledgments}
DL is supported by the Science \& Technology Development Fund of Tianjin Education Commission for Higher Education (2018KJ217), and the China Scholarship Council (201609345008). HL is supported by the Super post-doc program funded by Shanghai Municipal Government.

%\section*{References}
\bibliographystyle{model2-names}

\bibliography{ref}

\end{document}